\documentclass[letterpaper, 10 pt, conference]{ieeeconf}  

\usepackage{graphicx}
\usepackage{amsmath,amssymb}
\usepackage{booktabs}
\usepackage{cite}
\usepackage{xcolor}
\definecolor{myorange}{RGB}{237,125,49}
\definecolor{myred}{RGB}{255,0,0}

\makeatletter
\let\NAT@parse\undefined
\makeatother

\usepackage[hidelinks]{hyperref}

\IEEEoverridecommandlockouts                              

\title{\LARGE \bf
VDGS: Visibility-Driven Large-Scale 3D Gaussian Splatting for Aerial Scene Reconstruction
}

\author{Haolin Yu$^{1}$, Jiadong Tang$^{1}$, YiXian Wang$^{1}$, Yu Gao$^{1,*}$, Shi He$^{1}$, Zhilin Lai$^{2}$, Yi Yang$^{1}$, and Mengyin Fu$^{1}$ 
\thanks{This work was partly supported by the National Natural Science Foundation of China under Grant 62233002 and Grant 92370203.}%
\thanks{$^{1}$Haolin Yu, Jiadong Tang, YiXian Wang, Yu Gao, Shi He, Yi Yang, and Mengyin Fu are with the Beijing Institute of Technology, Beijing, China,
       $^{*}$Corresponding author: Yu Gao {\tt\small yugao\_it@163.com}}%
\thanks{$^{2}$Zhilin Lai is with Guangzhou Saite Intelligent Technology Co., Ltd.}%
}

\makeatletter
\def\@IEEEaftertitletext{%
  \refstepcounter{figure}%
  \centerline{\includegraphics[width=\textwidth]{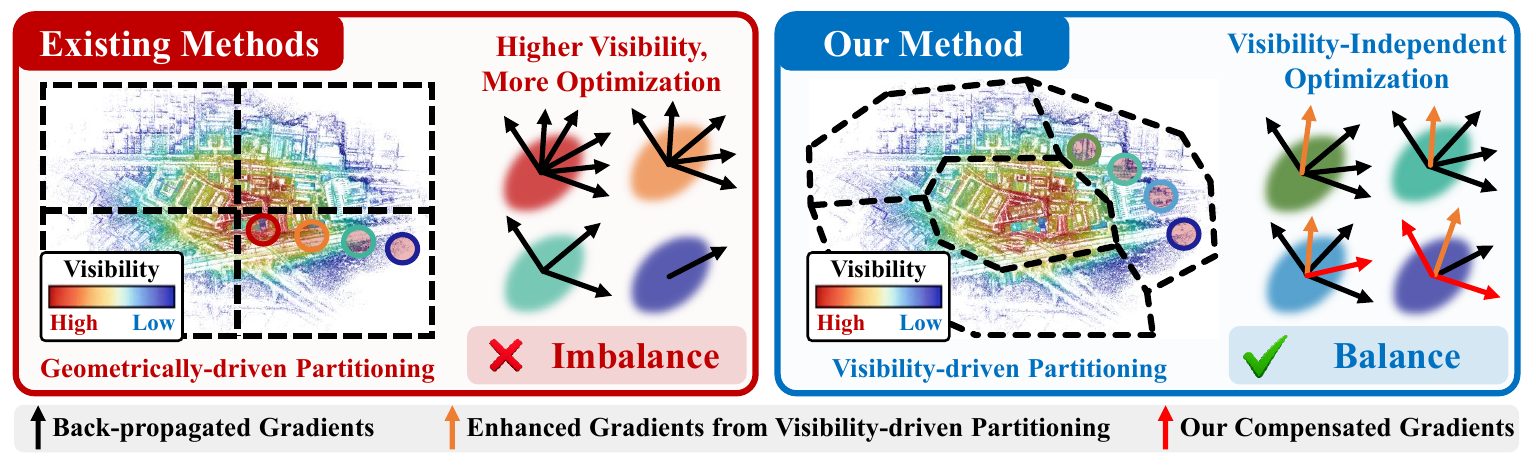}}%
  \vspace{0.5\baselineskip}%
  \centerline{%
    \parbox{\textwidth}{\footnotesize
    Fig.~\thefigure.
    \textbf{Left:} Existing divide-and-conquer strategies typically partition large-scale scenes primarily based on spatial geometry, while largely overlooking the interaction between camera distribution and scene structure. Consequently, individual blocks often suffer from highly uneven observation coverage, where back-propagated gradients \textbf{(black arrows)} are disproportionately biased toward highly visible regions, resulting in imbalanced optimization and degraded reconstruction quality.
    \textbf{Right:} In contrast, our VDGS framework adopts a visibility-driven partitioning strategy that encourages more balanced visibility distributions within each block. This leads to stronger and more evenly distributed gradient signals from weakly observed regions \textcolor{myorange}{\textbf{(orange arrows)}}. Furthermore, a dedicated gradient compensation mechanism is introduced to explicitly amplify these gradients \textcolor{myred}{\textbf{(red arrows)}}, jointly balancing optimization strength across regions and ultimately yielding improved reconstruction quality.
    }%
  }%
  \label{fig:teaser_compare}%
  \vspace{1\baselineskip}%
}
\makeatother

\begin{document}

\maketitle
\thispagestyle{empty}
\pagestyle{empty}


\begin{abstract}

Large-scale scene reconstruction is a critical foundational technology in robotic autonomous systems such as 3D mapping and autonomous driving. In recent years, 3D Gaussian Splatting (3DGS) has demonstrated remarkable advantages in both visual quality and computational efficiency, making it a promising representation for large-scale scene reconstruction. However, it still faces challenges in large-scale scenes, including excessive memory consumption and uneven viewpoint coverage caused by UAV acquisition, limiting its real-world applications. To address this, we propose VDGS, a novel 3DGS framework that incorporates camera distribution into scene modeling. VDGS introduces visibility-driven statistics for scene anchors to quantify supervision strength. These statistics are further leveraged for scene partitioning and for gradient compensation in under-optimized regions, thereby promoting balanced optimization across different regions. Extensive experiments on multiple large-scale aerial scene datasets demonstrate that, under imbalanced viewpoint distributions, VDGS consistently outperforms existing methods, while maintaining competitive performance in scenarios with more uniform view distributions.

\end{abstract}


\section{Introduction}

Large-scale scene reconstruction plays a fundamental role in robotics and autonomous systems, supporting critical applications such as 3D mapping \cite{predrecon,slam3r,long3r} and autonomous driving \cite{liu2020high,Vision,driv3r}. The emergence of NeRF \cite{NeRF} has provided a viable pathway toward high-fidelity large-scale 3D mapping; however, it is severely constrained by GPU memory limitations in large-scale scenes \cite{Mega-NeRF}. By replacing implicit MLP representations with an explicit Gaussian-based formulation, 3D Gaussian Splatting (3DGS) \cite{Kerbl3DGS} significantly improves rendering efficiency, yet still encounters similar memory challenges in large-scale scenarios.

Existing approaches fall into two directions. The first is a divide-and-conquer strategy \cite{VastGaussian, CityGaussian, DOGS, HUG}, inspired by NeRF-based methods \cite{Block-NeRF, Mega-NeRF}: large scenes are partitioned into independent blocks trained in parallel across GPUs and then merged. However, these partitioning schemes are largely geometry-driven and ignore the distribution of camera observations, so regions with different observation levels receive imbalanced supervision, causing pronounced quality degradation in sparsely observed areas.

\begin{figure}[t]
  \centering
  \includegraphics[width=\columnwidth]{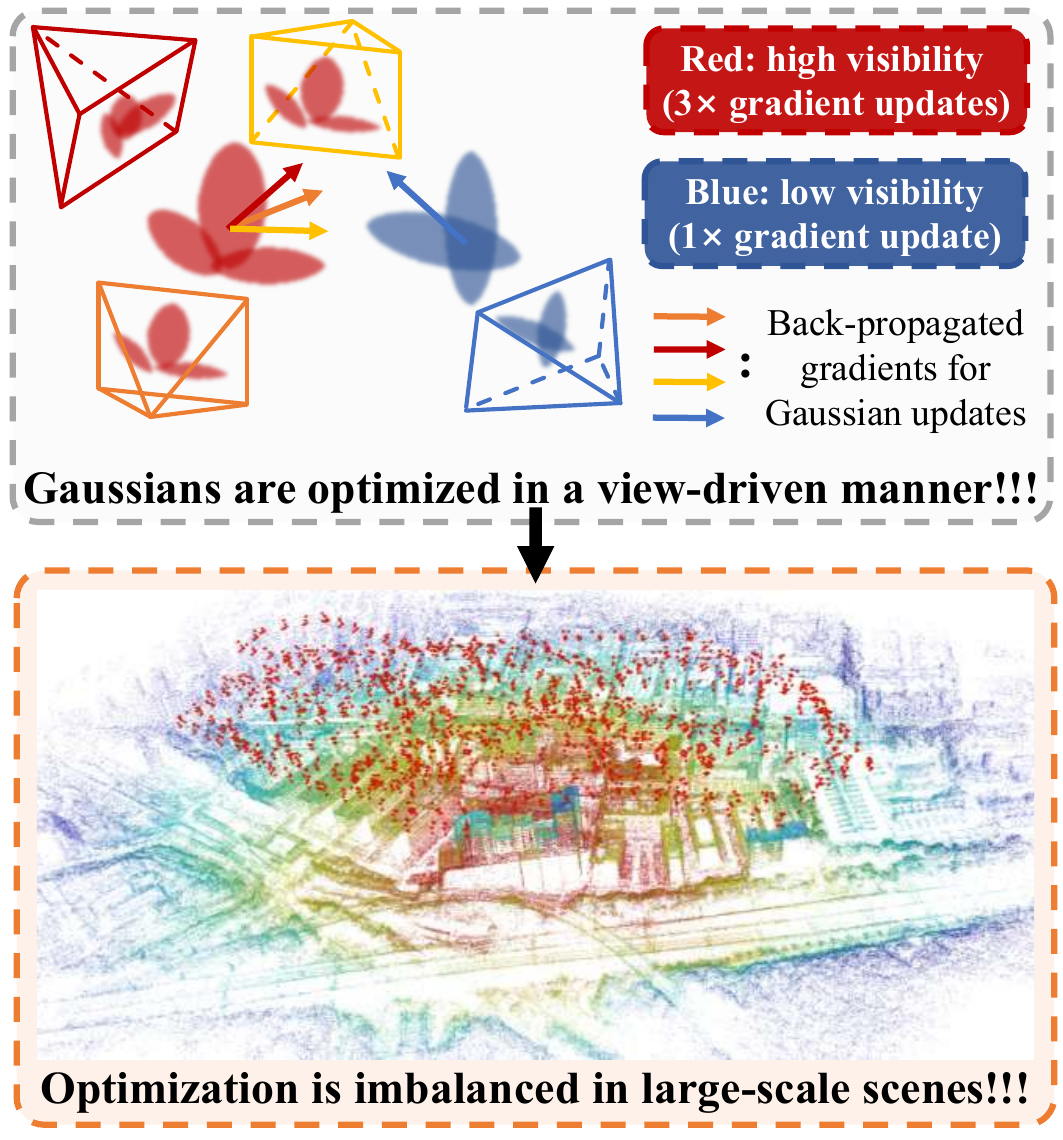}
  \caption{\textbf{Top}: 3DGS adopts a view-driven optimization scheme. The colored arrows associated with each Gaussian indicate gradient contributions from cameras of corresponding colors. Within one optimization cycle iterating over four camera views once, the red Gaussian receives three updates, whereas the blue Gaussian is updated only once.
  \textbf{Bottom}: Visualization of camera distribution and point visibility in the \textit{Residence} scene~\cite{UrbanScene3D}. The color of each point ranges from cool to warm, indicating low to high visibility. Point visibility serves as a proxy for the degree of optimization.}
  \label{fig:teaser_inherent}
\end{figure}

Another direction focuses on hybrid representations \cite{Scaffold-GS, Octree-GS, Momentum-GS, MixGS}, which combine implicit and explicit features to ease memory constraints while improving quality. These mostly adopt anchor-based structures over sparse 3D Gaussian fields, where Gaussian primitives are predicted from anchors via MLPs, making reconstruction quality highly dependent on the anchor distribution. Since anchor generation is governed by gradient-accumulation criteria, Gaussians in low-visibility regions receive weaker supervision and gradients, limiting their densification and thus the representational capacity there, as shown in Fig.~\ref{fig:teaser_inherent}.

To address these issues, we propose VDGS, a 3DGS framework that incorporates visibility into scene modeling. VDGS uses visibility-driven statistics to partition scene anchors into sub-blocks processed in parallel, and applies a gradient compensation strategy for anchors with insufficient viewpoint supervision. This alleviates supervision imbalance across regions and improves performance in low-visibility areas, yielding more stable training and robust reconstruction. We evaluate VDGS on challenging large-scale aerial datasets, including MatrixCity~\cite{Matrixcity}, UrbanScene3D~\cite{UrbanScene3D}, and Mill19~\cite{Mega-NeRF}. Under uneven viewpoint distributions, VDGS significantly outperforms state-of-the-art methods, while remaining competitive on uniform-view scenes, highlighting its robustness and generalization.

Existing large-scale 3DGS methods, such as VastGaussian~\cite{VastGaussian}, CityGaussian~\cite{CityGaussian}, and Momentum-GS~\cite{Momentum-GS}, do not explicitly model camera-guided strategies during scene partitioning and training. In contrast, VDGS incorporates visibility statistics into both scene partitioning and gradient-compensated anchor generation, directly addressing the optimization imbalance caused by uneven viewpoint coverage.

In summary, our main contributions are:
\begin{enumerate}
  \item We propose a Visibility-Driven Data Partitioning strategy that explicitly models the relationship between camera distribution and scene structure, enabling more principled data partitioning and more balanced training.
  \item We design a gradient compensation and densification mechanism tailored for low-visibility anchors, which effectively alleviates underfitting in weakly constrained regions and substantially improves reconstruction robustness in complex scenes.
  \item We conduct extensive experiments on multiple challenging real-world large-scale datasets, demonstrating that under uneven viewpoint coverage, the proposed method significantly outperforms existing approaches.
\end{enumerate}

\begin{figure*}[t]
  \centering
  \includegraphics[width=\textwidth]{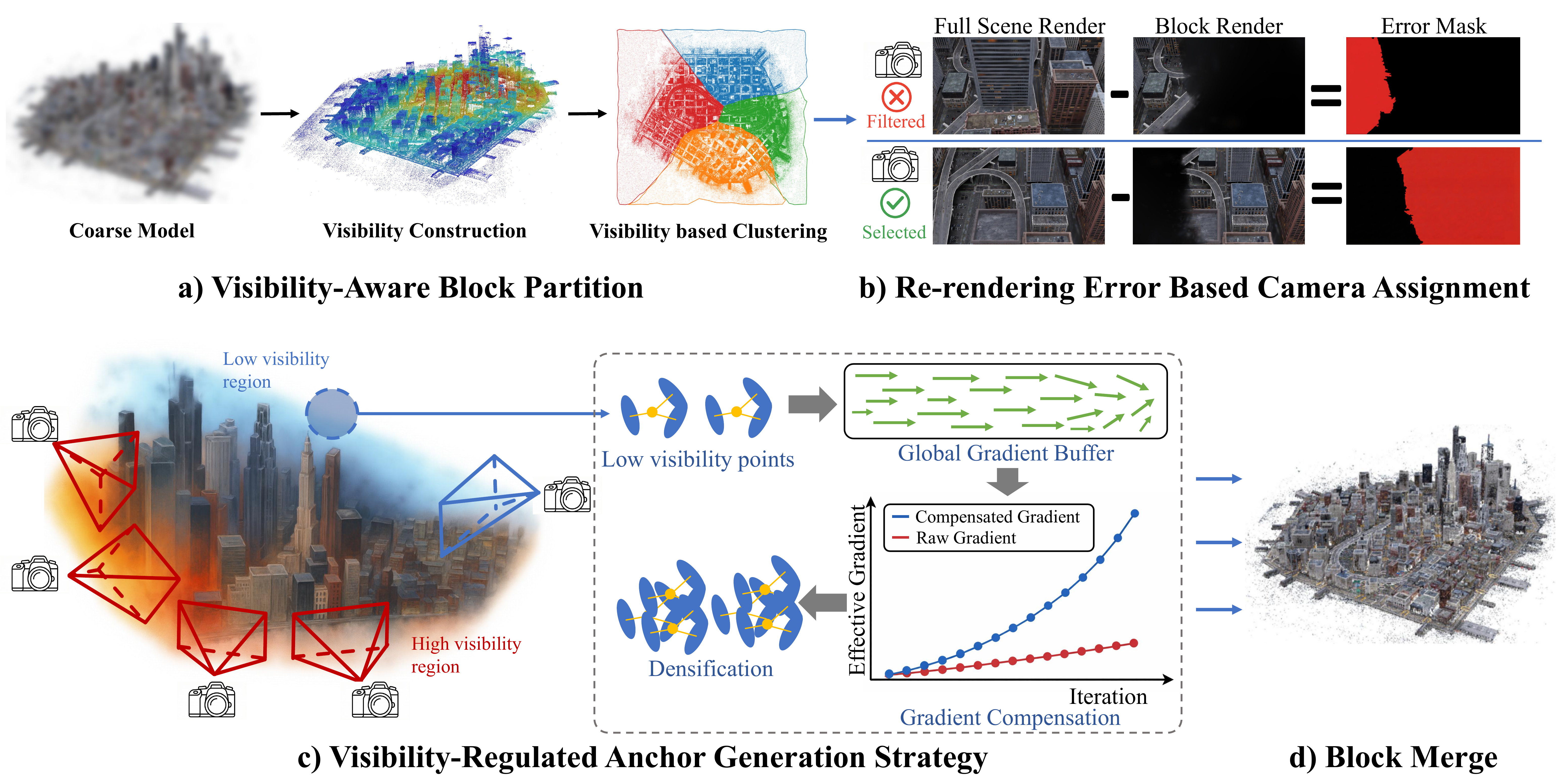}
  \caption{Overview of the proposed VDGS framework. 
  (a) \textbf{Visibility-Driven Data Partitioning.} The scene is partitioned into blocks using a visibility-aware clustering strategy, where clustering and partitioning are performed on planar primitives extracted from an initially trained coarse model.
  (b) \textbf{Re-rendering Error Based Camera Assignment.} A camera view is assigned to a block when the rendered area of that block in the view exceeds a predefined threshold. 
  (c) \textbf{Visibility-Guided Anchor Generation.} During training, gradients of low-visibility anchors are accumulated into a global buffer and later used for gradient compensation. This reinforcement alleviates weak supervision and promotes anchor densification in under-observed regions. 
  (d) \textbf{Block Merge.} After training, all blocks are merged according to visibility criteria to produce the final rendering result.}
  \label{fig:VDGS_overview}
\end{figure*}

\section{Related Work}

\subsection{Large Scene Reconstruction}

Large-scale scene reconstruction has been studied for decades. Traditional approaches rely on Structure-from-Motion (SfM) to estimate camera poses and sparse point clouds \cite{BuildingRoomInADay,PhotoTourism,ModelingAndRecognizition,DetailedUrban3D,SfM}. More recently, NeRF-based methods \cite{Block-NeRF,Mega-NeRF,Grid-NeRF,Switch-NeRF,switchnerfplus,bungeenerf} markedly improved realism by partitioning a scene into sub-regions and adopting a divide-and-conquer paradigm to ease computational and storage constraints, making divide-and-conquer a mainstream solution. For example, Block-NeRF \cite{Block-NeRF} improves robustness to illumination and viewpoint changes via appearance embeddings and learnable pose optimization, while Mega-NeRF \cite{Mega-NeRF} boosts training and inference efficiency through sparse voxel grids, underscoring the importance of principled data partitioning for scalable radiance field modeling.

The advent of 3D Gaussian Splatting (3DGS) \cite{Kerbl3DGS}, with its high-quality real-time rendering, opened a new avenue for large-scale reconstruction. To improve scalability, VastGaussian \cite{VastGaussian} and CityGaussian \cite{CityGaussian} alleviate memory pressure via block-wise training, while DoGaussian \cite{DOGS} adopts recursive partitioning with global consistency constraints. Level-of-detail methods such as H3DGS \cite{H3DGS}, Octree-GS \cite{Octree-GS}, and CityGaussianV2 \cite{CityGaussianV2} improve efficiency from a multi-resolution perspective, while HUG \cite{HUG}, Momentum-GS \cite{Momentum-GS}, and MixGS \cite{MixGS} further enhance quality through hierarchical image supervision, teacher–student distillation, and hybrid coarse-Gaussian representations, respectively. Despite steady gains, these methods focus on geometric partitioning, load balancing, or hierarchical representations, and largely ignore training constraints induced by camera frustum distributions, still suffering from unstable reconstruction in scenes with uneven camera coverage.

\subsection{Novel View Synthesis}

NeRF \cite{NeRF} and 3D Gaussian Splatting (3DGS) \cite{Kerbl3DGS} have driven rapid progress in novel view synthesis. NeRF represents scenes implicitly via coordinate-based MLPs, achieving high-fidelity rendering but at high training and rendering cost. To improve efficiency, Instant-NGP \cite{Instant-NGP} uses multi-resolution hash encoding and DVGO \cite{DVGO} explicit voxel grids, while PyNeRF \cite{PyNeRF}, Neuralangelo \cite{Neuralangelo}, and Binary Opacity Grids \cite{Binary} enhance detail and geometric consistency through multi-scale modeling and explicit structure.

In contrast, 3DGS \cite{Kerbl3DGS} explicitly models scenes with anisotropic Gaussian ellipsoids, enabling real-time, high-quality rendering. Following this line, 2DGS \cite{Huang2DGS} improves geometric accuracy, while Scaffold-GS \cite{Scaffold-GS} introduces sparse anchors and neural decoding to reduce memory and improve robustness to viewpoint changes. Nevertheless, these methods generally assume spatially uniform supervision and lack explicit modeling of the optimization imbalance caused by camera distribution and visibility variations, limiting their robustness in large-scale aerial scenes with highly uneven viewpoint coverage.


\section{Preliminaries}

\subsection{3DGS}

3DGS represents a scene with a collection of anisotropic 3D Gaussian primitives, each parameterized by a center $\boldsymbol{\mu} \in \mathbb{R}^3$ and a covariance matrix $\boldsymbol{\Sigma} \in \mathbb{R}^{3 \times 3}$, defining the spatial distribution
\begin{equation}
G(x) = \exp\left(-\frac{1}{2}(x - \boldsymbol{\mu})^\top \boldsymbol{\Sigma}^{-1}(x - \boldsymbol{\mu})\right).
\end{equation}

Each Gaussian carries an opacity $\sigma$ and a view-dependent color $\mathbf{c}$ from spherical harmonics. Gaussians are projected onto the image plane and composited front-to-back via $\alpha$-blending, giving the color at pixel $x'$:
\begin{equation}
C(x') = \sum_{i \in \mathcal{N}} T_i \, \mathbf{c}_i \, \sigma_i, 
\qquad
T_i = \prod_{j=1}^{i-1} (1 - \sigma_j),
\end{equation}
where $\mathcal{N}$ denotes Gaussians contributing to pixel $x'$.

\subsection{Scaffold-GS}

Scaffold-GS introduces an anchor-based structured representation to improve 3DGS scalability for large scenes. Each anchor at position $\mathbf{x}_v$ holds a learnable feature $\mathbf{f}_v$, from which $k$ neural Gaussians are decoded with centers given by predicted relative offsets:
\begin{equation}
\{\boldsymbol{\mu}_0, \ldots, \boldsymbol{\mu}_{k-1}\}
= \mathbf{x}_v + \{\mathbf{O}_0, \ldots, \mathbf{O}_{k-1}\} \cdot l_v ,
\end{equation}
where $\mathbf{O}_i$ denotes predicted offsets and $l_v$ sets magnitude.

Other attributes (opacity, scale, rotation, color) are decoded from the anchor feature via lightweight MLPs, e.g., opacity:
\begin{equation}
\{\alpha_0, \ldots, \alpha_{k-1}\} = F_{\alpha}(\mathbf{f}_v, \Delta v_c, \mathbf{d}_{vc}),
\end{equation}
where $\Delta v_c$ and $\mathbf{d}_{vc}$ encode camera-relative geometry and viewing direction. The neural Gaussians are rendered with the standard 3DGS rasterizer, and anchor-level densification and pruning maintain efficiency during training.

\section{Methods}

\subsection{Overview} 
Anchor-based representations implicitly encode geometry and appearance into compact anchor parameters, enabling memory-efficient modeling for large-scale aerial reconstruction. However, in city-scale scenes, they suffer from optimization imbalance caused by uneven camera-induced visibility. To address this issue, we propose a visibility-aware anchor-based framework that explicitly incorporates regional visibility into the training pipeline. As illustrated in Fig.~\ref{fig:VDGS_overview}, the framework consists of a block-wise training strategy for Visibility-Driven Data Partitioning (Section B) and a Visibility-Guided Anchor Generation scheme for balanced optimization within each block (Section C).

\begin{figure}[t]
  \centering
  \includegraphics[width=\columnwidth]{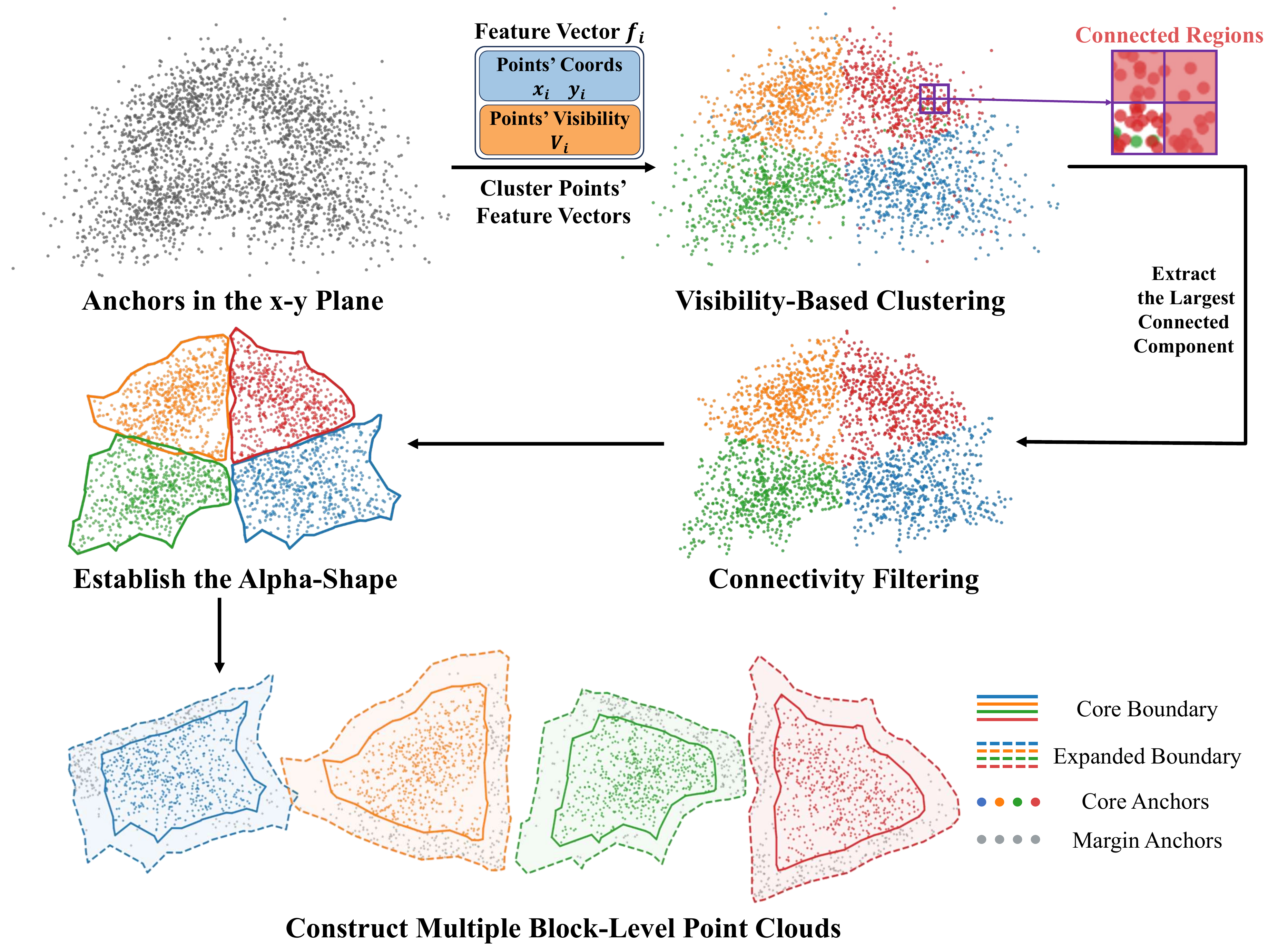}
  \caption{\textbf{Partition pipeline.} 
    First, feature vectors are constructed for anchors using planar $x$--$y$ coordinates and visibility statistics, and anchors are clustered in the resulting feature space. 
    Next, connectivity filtering is performed within each cluster to retain the largest connected component, based on which an $\alpha$-shape defines the block boundary. 
    Finally, the block boundaries are further expanded to obtain the final block-level partitions.}
  \label{fig:Partition_Pipeline}
\end{figure}

\subsection{Visibility-Driven Data Partitioning} 
A salient characteristic of large-scale aerial scenes is their vast spatial extent combined with relatively sparse visibility distributions, leading to substantial variation in viewpoint overlap across regions and, consequently, pronounced imbalance in regional visibility. However, existing large-scale reconstruction methods typically rely on geometric partitioning \cite{VastGaussian}, regular grid-based division \cite{HUG}, or computational load balancing \cite{DOGS} when segmenting scenes, and rarely explicitly account for such visibility differences across regions. As a result, certain regions may be optimized under limited multi-view constraints, which ultimately degrades overall reconstruction quality. Motivated by these observations, we propose a Visibility-Driven Data Partitioning and camera assignment framework that explicitly incorporates scene visibility during partitioning, thereby alleviating optimization imbalance caused by regional visibility.

\paragraph{Visibility-Driven Block Partition}
To improve robustness across scenes, we design a hybrid partitioning strategy that jointly accounts for spatial location and anchor visibility, as illustrated in Fig.~\ref{fig:Partition_Pipeline}.

Firstly, following common practice in recent large-scale 3DGS methods \cite{CityGaussian,Momentum-GS}, we train the model on the full scene for a fixed number of iterations to obtain a coarse global model. This stage provides a global prior with basic geometry, appearance consistency, and reliable estimates of camera--block relevance.
Next, we combine the normalized planar anchor coordinates on the x–y plane, $(\hat{x}_i, \hat{y}_i)$, with visibility statistic $V_i$ to form a clustering feature:
\begin{equation}
f_i =
\begin{bmatrix}
\hat{x}_i \\[2pt]
\hat{y}_i \\[2pt]
\lambda_{\text{vis}}\,\hat{V}_i
\end{bmatrix},
\qquad
\hat{V}_i = \frac{V_i - V_{\min}}{V_{\max} - V_{\min} + 10^{-8}} .
\end{equation}

Here, $(\hat{x}_i, \hat{y}_i)$ denote the normalized anchor coordinates on the $x$–$y$ plane.
$V_i$ represents the visibility of anchor $i$, defined as the number of training camera frustums that observe it, while $V_{\min}$ and $V_{\max}$ are the minimum and maximum visibility values among all anchors, respectively.
$\hat{V}_i$ denotes the normalized visibility, and $\lambda_{\text{vis}}$ is a weighting factor that controls the influence of visibility in the feature space.

We cluster anchors in the joint spatial--visibility feature space to form block-level partitions, such that anchors within each block exhibit relatively consistent visibility levels, thereby alleviating visibility imbalance during optimization.

To obtain spatially coherent blocks, we retain the largest connected component within each cluster and derive its 2D boundary using $\alpha$-shape construction \cite{alphashape}. The resulting boundaries are further expanded to ensure cross-block consistency and define the final spatial extent of each block.

\paragraph{Re-rendering Error Based Camera Assignment}
After scene partitioning, we assign training viewpoints to each block by following a re-rendering error criterion, enabling precise and adaptive camera batching for block-wise training.

Specifically, for each candidate camera view, we render both the full-scene image $I^{\text{full}}$ and the block-level image $I^{\text{block}}$, and compute a pixel-wise difference map as
\begin{equation}
\Delta(\mathbf{u})
= \tfrac{1}{3} \sum_{c=1}^3 \big| I^{\text{block}}_c(\mathbf{u}) - I^{\text{full}}_c(\mathbf{u}) \big| ,
\end{equation}
where $\mathbf{u}$ denotes pixel locations and $c$ indexes color channels.
Based on this, we define the normalized coverage ratio
\begin{equation}
\bar{\Delta}
= \frac{1}{|\Omega|} \sum_{\mathbf{u} \in \Omega} \mathbb{I}\!\left( \Delta(\mathbf{u}) < \tau \right),
\end{equation}
where $\Omega$ denotes the image domain, $|\Omega|$ is the total number of pixels, and $\tau$ is a predefined pixel-wise difference threshold.

If $\bar{\Delta}$ exceeds a camera assignment threshold, the corresponding camera is assigned to the block for subsequent training. As this criterion is directly derived from re-rendering error, it robustly characterizes camera--block relationships even under occlusions or complex viewpoint overlaps, and generalizes well across diverse scenes.

\paragraph{Training and Block Merging}

After data partitioning, the sub-blocks are distributed across multiple GPUs and trained in parallel. To maintain cross-block consistency while enabling independent block-wise optimization, we suppress gradient updates at block boundaries during training and simultaneously freeze the parameters of the MLP decoder. This design prevents unstable gradients near block boundaries from interfering with local reconstruction quality.

Once sub-block training is completed, we merge core anchors from each block based on visibility coverage. This training and merging pipeline yields a complete, coherent, and globally consistent scene representation.

\begin{table*}[t]
\centering
\caption{Quantitative comparison of our \textbf{VDGS} with prior methods across five large-scale aerial scenes.
We report PSNR ($\uparrow$), SSIM ($\uparrow$), and LPIPS ($\downarrow$) on test views.
The \textbf{best} and \underline{second-best} scores are highlighted.}
\label{tab:quantitative}

\setlength{\tabcolsep}{5pt}
\renewcommand{\arraystretch}{1.05}

\resizebox{\textwidth}{!}{
\begin{tabular}{l ccc ccc ccc ccc ccc}
\toprule
\textbf{Method}
& \multicolumn{3}{c}{\textit{Residence}}
& \multicolumn{3}{c}{\textit{Sci-Art}}
& \multicolumn{3}{c}{\textit{Rubble}}
& \multicolumn{3}{c}{\textit{Building}}
& \multicolumn{3}{c}{\textit{Block-All}} \\
\cmidrule(lr){2-4} \cmidrule(lr){5-7} \cmidrule(lr){8-10}
\cmidrule(lr){11-13} \cmidrule(lr){14-16}

& PSNR$\uparrow$ & SSIM$\uparrow$ & LPIPS$\downarrow$
& PSNR$\uparrow$ & SSIM$\uparrow$ & LPIPS$\downarrow$
& PSNR$\uparrow$ & SSIM$\uparrow$ & LPIPS$\downarrow$
& PSNR$\uparrow$ & SSIM$\uparrow$ & LPIPS$\downarrow$
& PSNR$\uparrow$ & SSIM$\uparrow$ & LPIPS$\downarrow$ \\
\midrule

Switch-NeRF \cite{Switch-NeRF}
& 22.57 & 0.654 & 0.352
& \underline{26.51} & 0.795 & 0.271
& 24.31 & 0.562 & 0.478
& 21.54 & 0.579 & 0.397
& -- & -- & -- \\

3DGS \cite{Kerbl3DGS}
& 21.94 & 0.764 & 0.279
& 21.85 & 0.787 & 0.311
& 25.20 & 0.757 & 0.318
& 22.04 & 0.728 & 0.332
& 27.36 & 0.818 & 0.237 \\

Scaffold-GS \cite{Scaffold-GS}
& 22.00 & 0.761 & 0.286
& 22.56 & 0.796 & 0.302
& 24.83 & 0.721 & 0.353
& 22.42 & 0.719 & 0.335
& 27.13 & 0.868 & 0.210 \\

DOGS \cite{DOGS}
& 21.94 & 0.740 & 0.244
& 24.42 & 0.804 & 0.219
& 25.78 & 0.765 & 0.257
& 22.73 & 0.759 & 0.214
& 28.58 & 0.847 & 0.219 \\

Octree-GS \cite{Octree-GS}
& 22.29 & 0.762 & 0.288
& 23.38 & 0.828 & 0.240
& 25.34 & 0.763 & 0.299
& 23.66 & 0.776 & 0.267
& 26.41 & 0.814 & 0.282 \\

HUG \cite{HUG}
& 22.33 & 0.813 & 0.207
& 21.83 & 0.846 & \underline{0.204}
& 26.42 & \underline{0.839} & \underline{0.197}
& 22.35 & 0.792 & 0.228
& 28.02 & \textbf{0.883} & \textbf{0.142} \\

Momentum-GS \cite{Momentum-GS}
& 23.37 & \underline{0.828} & \underline{0.196}
& 25.06 & \underline{0.860} & \underline{0.204}
& \underline{26.66} & 0.826 & 0.200
& \underline{23.65} & \textbf{0.813} & \textbf{0.194}
& \textbf{29.11} & \underline{0.881} & \underline{0.180} \\

MixGS \cite{MixGS}
& \underline{23.39} & 0.815 & 0.219
& 24.20 & 0.856 & 0.220
& \underline{26.66} & 0.792 & 0.267
& 23.03 & 0.771 & 0.261
& -- & -- & -- \\

\midrule
\textbf{VDGS (Ours)}
& \textbf{24.03} & \textbf{0.845} & \textbf{0.188}
& \textbf{26.94} & \textbf{0.870} & \textbf{0.180}
& \textbf{26.67} & \textbf{0.852} & \textbf{0.188}
& \textbf{23.73} & \underline{0.800} & \underline{0.208}
& \underline{28.74} & 0.863 & 0.204 \\

\bottomrule
\end{tabular}
}
\end{table*}

\begin{figure*}[t]
  \centering
  \includegraphics[width=\textwidth]{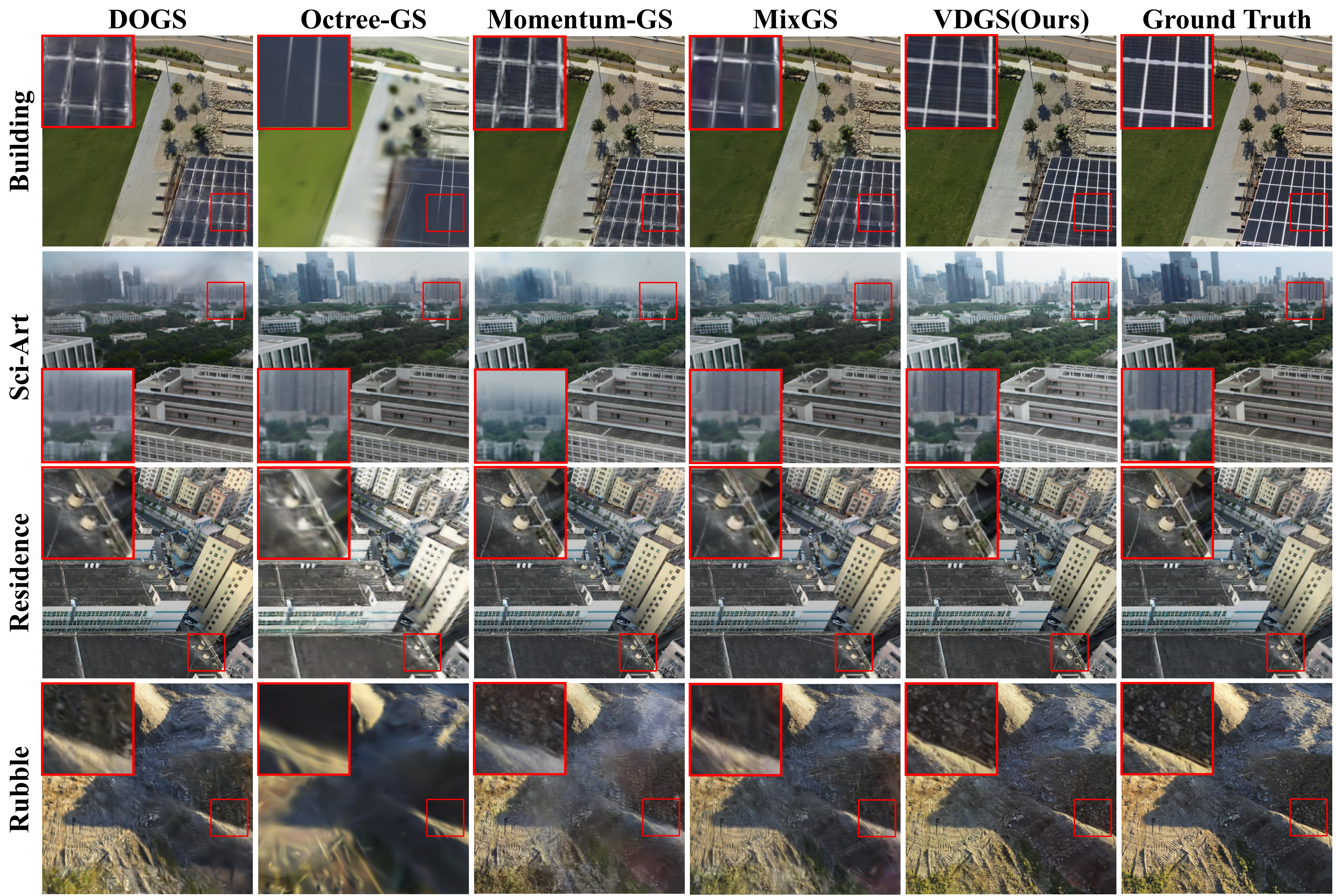}
  \caption{Qualitative rendering comparisons on the Mill19~\cite{Mega-NeRF} and UrbanScene3D~\cite{UrbanScene3D} datasets. Zoomed-in views of the regions highlighted by red boxes are shown for detailed inspection.}
  \label{fig:Quantitative_Compare}
\end{figure*}

\subsection{Visibility-Guided Anchor Generation}

For large-scale scene anchor representations, anchors constructed solely from the initial point cloud are often insufficient to capture complex structures. To address this, existing anchor-based methods generally follow the vanilla 3DGS paradigm, generating new anchors from accumulated gradients backpropagated from the rendering loss. However, large-scale aerial scenes typically exhibit highly uneven viewpoint coverage and pronounced regional optimization variation, resulting in severely imbalanced gradient statistics across space. Consequently, anchor densification probabilities vary significantly across regions, leaving some areas under-optimized with too few anchors.

Since the reconstruction quality of anchor-based methods is highly dependent on the spatial distribution and density of anchors, such gradient imbalance further amplifies optimization disparities across regions, ultimately leading to pronounced spatial inconsistency in reconstruction quality for large-scale aerial scenes.

To mitigate anchor optimization bias induced by visibility differences, we propose a Visibility-Guided Anchor Generation strategy. By explicitly incorporating visibility information to guide anchor densification decisions, the proposed method promotes more balanced gradient updates across regions during training, effectively alleviating uneven anchor distribution in large-scale aerial scenes.

\paragraph{Constructing Low-Visibility Masks}
Anchor densification is triggered periodically at fixed training intervals, during which new anchors are placed in unoccupied voxels that exhibit high accumulated gradients.

We define a cycle-wise variable $c_k \in [0,1]$, updated at each densification period, recording the fraction of cameras that observe the $k$-th Gaussian within the current cycle:
\begin{equation}
c_k = \frac{1}{V} \sum_{v=1}^{V} \mathbb{I}\!\left(\mathbf{x}_{a \rightarrow k} \in \operatorname{Frustum}(v)\right) ,
\end{equation}
where $\mathbb{I}(\cdot)$ denotes the indicator function, $\mathbf{x}_{a \rightarrow k}$ represents the center of the source anchor $a$ that generates Gaussian $k$, $V$ is the total number of cameras in the current cycle, and $\operatorname{Frustum}(v)$ is the viewing frustum of the $v$-th camera.

Based on this observation ratio, we define the set of low-visibility Gaussians using a visibility threshold $T_{\text{low}} \in (0,1)$, i.e., Gaussians observed by fewer than a fraction $T_{\text{low}}$ of the cameras in the current cycle:
\begin{equation}
\mathcal{L} = \{\, k \mid c_k < T_{\text{low}} \,\} .
\end{equation}
The set $\mathcal{L}$ collects all Gaussians insufficiently observed in the current cycle, which are then passed to the gradient compensation module to mitigate viewpoint-induced bias in gradient statistics.

\paragraph{Gradient Cache with Exponential Compensation Mechanism}
To counter the effect of uneven viewpoints on anchor densification, we apply gradient compensation to the identified low-visibility Gaussians, maintaining for each such Gaussian $k$ a global buffer $g_k$ that accumulates its historical gradients.

At the end of each densification cycle, the gradient cache is updated as
\begin{equation}
g_k \leftarrow \gamma\, g_k^{\text{old}} + \beta_g \, \bar{g}_k ,
\end{equation}
where $\gamma$ is a decay factor that limits long-term over-correction, $\beta_g$ scales the visibility-aware adjustment, and $\bar{g}_k$ is the average gradient magnitude of Gaussian $k$ in the current cycle.

During densification, $g_k$ is added to the raw gradient statistics of the corresponding Gaussian. Thus, even when low-visibility regions receive limited direct gradients within a cycle, accumulated historical contributions still yield reasonable densification opportunities, alleviating anchor-generation bias caused by uneven viewpoint coverage.

\section{Experiments}

\subsection{Experimental Setup}

\paragraph{Datasets and Metrics} We conduct experiments on five large-scale UAV-view datasets: the \textit{Residence} and \textit{Sci-Art} scenes from UrbanScene3D \cite{UrbanScene3D}, the \textit{Building} and \textit{Rubble} scenes from the Mill19 dataset \cite{Mega-NeRF}, and the \textit{Block-All} scene from MatrixCity \cite{Matrixcity}. Each dataset contains thousands of high-resolution images. Following prior work \cite{Mega-NeRF,VastGaussian,MixGS}, we downsample images by a factor of four for all datasets except MatrixCity, whose width is resized to 1600 pixels, and adopt the same train/test splits as previous methods \cite{Mega-NeRF,CityGaussian}. Reconstruction quality is evaluated using PSNR, SSIM \cite{SSIM}, and LPIPS \cite{LPIPS}.


\paragraph{Implementation Details and Compared Methods}
We use camera poses and sparse point clouds estimated by COLMAP \cite{SfM} as input.
Training starts with an initial coarse stage of 100{,}000 iterations to obtain a global model, followed by 50{,}000 iterations for each sub-block.
For all scenes, the scene is uniformly partitioned into eight blocks.
In the block partitioning strategy, $\lambda_{\text{vis}}$ is set within $[0.7, 1.3]$, with a boundary expansion ratio of $0.4$.
For camera assignment, we use a view selection threshold where the pixel coverage ratio $\bar{\Delta}$ exceeds $0.45$.
For anchor generation, we set $T_{\text{low}} = 0.20$, $\gamma = 0.9$, and $\beta_g = 2.0$.
We apply the same color correction strategy as prior work \cite{DOGS,Momentum-GS} when computing evaluation metrics.
We compare our method with Switch-NeRF \cite{Switch-NeRF}, 3DGS \cite{Kerbl3DGS}, Scaffold-GS \cite{Scaffold-GS}, DOGS \cite{DOGS}, Octree-GS \cite{Octree-GS}, HUG \cite{HUG}, Momentum-GS \cite{Momentum-GS}, and MixGS \cite{MixGS}.
All experiments are conducted on an NVIDIA A100 GPU with 40\,GB of memory.

\begin{figure}[t]
  \centering
  \includegraphics[width=\columnwidth]{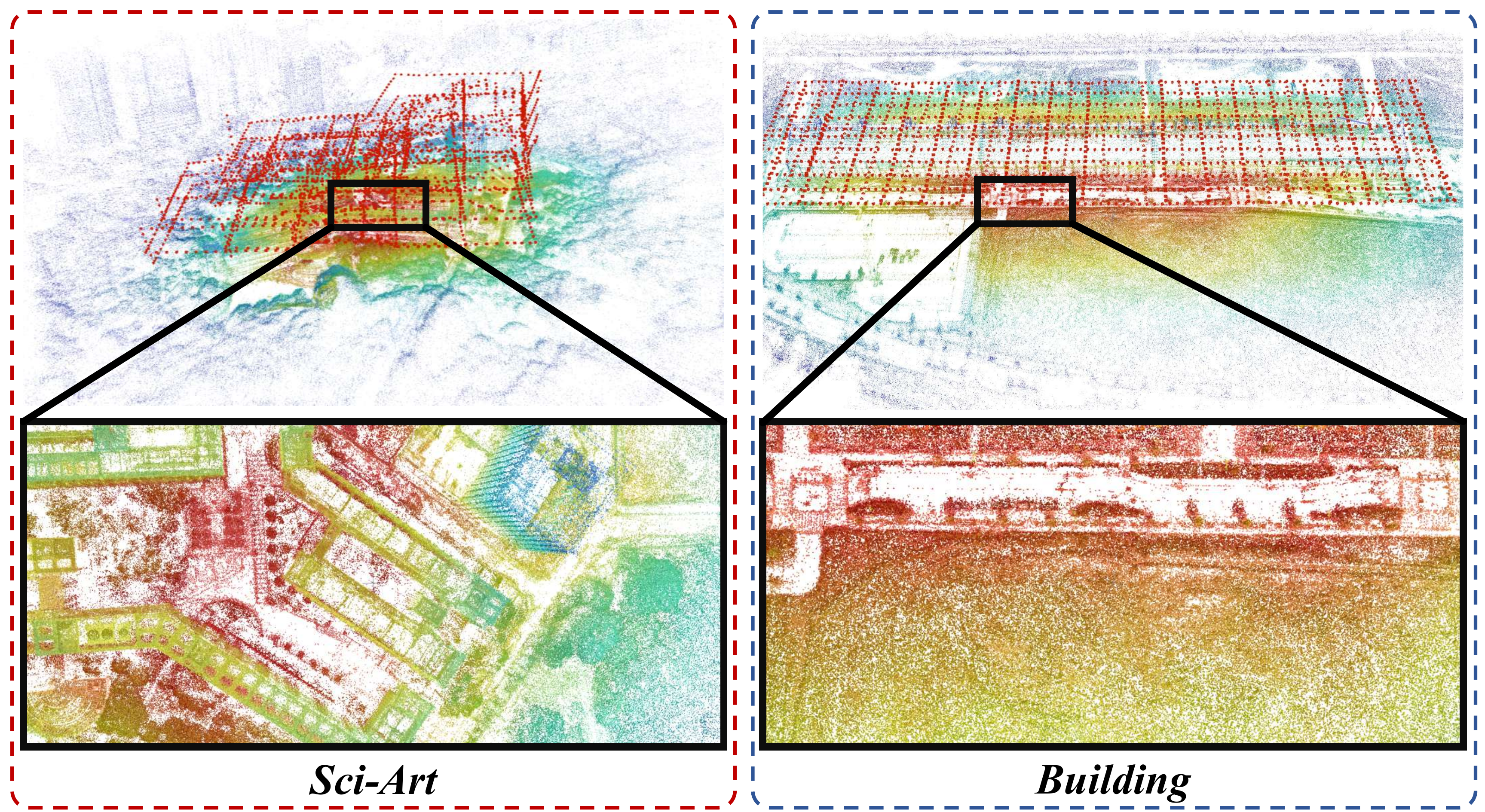}
  \caption{Camera and point visibility distributions for the \textit{Building} and \textit{Sci-Art} scenes. Camera frustums are red; warmer point colors denote higher visibility. \textit{Building} shows relatively uniform camera placement and visibility, whereas \textit{Sci-Art} exhibits more uneven coverage and less consistent visibility.}
  \label{fig:Scenes_visibility}
\end{figure}

\subsection{Results Analysis}

Table~\ref{tab:quantitative} reports quantitative results on five large-scale aerial scenes, and Fig.~\ref{fig:Scenes_visibility} illustrates visibility distributions for two representative ones. \textit{MatrixCity}, \textit{Building}, and \textit{Rubble} exhibit relatively uniform camera coverage, whereas \textit{Residence} and \textit{Sci-Art} suffer from pronounced viewpoint imbalance.
As shown both quantitatively (Table~\ref{tab:quantitative}) and qualitatively (Fig.~\ref{fig:Quantitative_Compare}), VDGS significantly outperforms existing methods in scenes with uneven visibility, as its Visibility-Driven Data Partitioning and Visibility-Guided Anchor Generation more effectively alleviate the optimization imbalance induced by extreme visibility and biased viewpoints.
It also remains competitive on more uniform scenes, indicating our strategy is not limited to specific configurations.
Notably, on \textit{Sci-Art}, VDGS achieves substantially higher PSNR than existing 3DGS-based methods. As shown in Fig.~\ref{fig:Scenes_visibility}, this scene contains extensive distant low-visibility regions with sparse viewpoint coverage; our Visibility-Guided Anchor Generation concentrates optimization and gradient compensation there, mitigating the imbalance caused by uneven camera distribution.

\begin{table}[h]
\centering
\caption{Ablation study on different training strategies evaluated on the Residence scene.
We report PSNR ($\uparrow$), SSIM ($\uparrow$), and LPIPS ($\downarrow$).}
\label{tab:Ablation1}
\setlength{\tabcolsep}{6pt}
\renewcommand{\arraystretch}{1.05}
\begin{tabular}{lccc}
\toprule
\textbf{Training Strategy} & \textbf{PSNR} $\uparrow$ & \textbf{SSIM} $\uparrow$ & \textbf{LPIPS} $\downarrow$ \\
\midrule
(a) Baseline      & 22.64 & 0.801 & 0.233 \\
(b) w/o Blocking  & 23.39 & 0.827 & 0.209 \\
(c) w/o Anchor Generation & 23.56 & 0.833 & 0.202 \\
(d) Full          & \textbf{24.01} & \textbf{0.845} & \textbf{0.188} \\
\bottomrule
\end{tabular}
\end{table}

\begin{table}[h]
\centering
\caption{Ablation study on the number of blocks evaluated on the Residence scene.
We report PSNR ($\uparrow$), SSIM ($\uparrow$), and LPIPS ($\downarrow$).}
\label{tab:Ablation2}
\setlength{\tabcolsep}{6pt}
\renewcommand{\arraystretch}{1.05}
\begin{tabular}{c c c c}
\toprule
\textbf{\#Blocks} & \textbf{PSNR} $\uparrow$ & \textbf{SSIM} $\uparrow$ & \textbf{LPIPS} $\downarrow$ \\
\midrule
1  & 23.39 & 0.827 & 0.209 \\
2  & 23.63 & 0.835 & 0.196 \\
4  & 24.01 & 0.845 & 0.188 \\
8  & 24.04 & \textbf{0.846} & \textbf{0.187} \\
16 & \textbf{24.05} & \textbf{0.846} & \textbf{0.187} \\
\bottomrule
\end{tabular}
\end{table}

\begin{table}[h]
\centering
\caption{Ablation on key hyperparameters evaluated on the \textit{Residence} scene. Each parameter is varied independently while keeping the others at their defaults ($\lambda_{\text{vis}}{=}1.0$, $T_{\text{low}}{=}0.20$, $\beta_g{=}2.0$, $\gamma{=}0.90$). PSNR\,($\uparrow$), SSIM\,($\uparrow$), LPIPS\,($\downarrow$).}
\label{tab:hyperparam}
\setlength{\tabcolsep}{4pt}
\renewcommand{\arraystretch}{1.0}

\begin{tabular}{lcccc}
\toprule
\textbf{Parameter / Value} & \textbf{PSNR}$\uparrow$ & \textbf{SSIM}$\uparrow$ & \textbf{LPIPS}$\downarrow$ \\
\midrule
\multicolumn{4}{l}{\textit{(a) Visibility weight $\lambda_{\text{vis}}$}} \\
\quad $\lambda_{\text{vis}}=0$ (spatial only) & 23.51 & 0.8353 & 0.1988 \\
\quad $\lambda_{\text{vis}}=0.5$ & 23.99 & 0.8433 & 0.1918 \\
\quad $\lambda_{\text{vis}}=1.0$ (default) & \textbf{24.03} & \textbf{0.8453} & \textbf{0.1884} \\
\quad $\lambda_{\text{vis}}=1.3$ & 23.80 & 0.8381 & 0.1928 \\
\midrule
\multicolumn{4}{l}{\textit{(b) Low-visibility threshold $T_{\text{low}}$}} \\
\quad $T_{\text{low}}=0.05$ & 23.72 & 0.8357 & 0.1989 \\
\quad $T_{\text{low}}=0.15$ & 23.94 & 0.8429 & 0.1913 \\
\quad $T_{\text{low}}=0.20$ (default) & \textbf{24.03} & \textbf{0.8453} & \textbf{0.1884} \\
\quad $T_{\text{low}}=0.30$ & 24.06 & 0.8456 & 0.1883 \\
\midrule
\multicolumn{4}{l}{\textit{(c) Compensation scale $\beta_g$}} \\
\quad $\beta_g=0.5$ & 23.88 & 0.8471 & 0.1927 \\
\quad $\beta_g=1.5$ & 23.92 & 0.8421 & 0.1906 \\
\quad $\beta_g=2.0$ (default) & \textbf{24.03} & \textbf{0.8453} & \textbf{0.1884} \\
\quad $\beta_g=3.0$ & 24.08 & 0.8465 & 0.1880 \\
\midrule
\multicolumn{4}{l}{\textit{(d) Decay factor $\gamma$}} \\
\quad $\gamma=0.80$ & 23.98 & 0.8445 & 0.1885 \\
\quad $\gamma=0.90$ (default) & \textbf{24.03} & \textbf{0.8453} & \textbf{0.1884} \\
\quad $\gamma=0.95$ & 23.94 & 0.8443 & 0.1888 \\
\bottomrule
\end{tabular}

\end{table}

\subsection{Ablation Studies}

\paragraph{Training Strategies} Table~\ref{tab:Ablation1} ablates the key components of VDGS on the Residence scene by removing individual modules. Removing the blocking strategy (\emph{w/o Blocking}) drops PSNR by 0.62~dB relative to the full model, showing that visibility-aware partitioning is critical to reconstruction quality. Removing the Visibility-Guided Anchor Generation (\emph{w/o Anchor Generation}) also degrades performance, highlighting the necessity of gradient compensation for anchors in low-visibility regions.

\paragraph{Partition-robustness} Table~\ref{tab:Ablation2} reports results under different numbers of partitions on the Residence scene. Once the partition count exceeds a certain threshold, all metrics stabilize and gains saturate, indicating that VDGS is highly robust to the partition number and maintains reliable quality across configurations.

\paragraph{Hyperparameter Sensitivity} Table~\ref{tab:hyperparam} reports sensitivity to the four main hyperparameters of VDGS. Across all tested ranges, PSNR, SSIM, and LPIPS remain stable, confirming that VDGS is not sensitive to precise hyperparameter choices. Notably, setting $\lambda_{\text{vis}}{=}0$ reduces partitioning to a purely spatial scheme, and the resulting performance drop confirms the effectiveness of our visibility-based partitioning strategy.

\paragraph{Efficiency Analysis} Taking the 8-block partition of the \textit{Residence} scene as an example, VDGS requires only \textbf{1.48~h wall-clock time} (11.85~GPU-hours) for the block training stage using 8 GPUs in parallel, with a peak per-block GPU memory of $\approx$5.4~GB, a merged model size of 847~MB, and a rendering speed of 42.96~FPS. Wall-clock time and total GPU-hours are reported separately because block training is parallelized, making GPU-hours a fairer measure of total compute cost. The 100{,}000-iteration coarse model training requires approximately 5~hours.

\section{Discussion}
Although VDGS is evaluated on aerial datasets, the pipeline is in principle applicable to other capture settings with uneven viewpoint coverage (e.g., street-level scenes). However, street-level scenes do not enjoy the relatively uniform viewpoint coverage characteristic of aerial capture, and indiscriminately compensating gradients in low-visibility regions may introduce unforeseen issues; definitive conclusions would require further experimentation.

\section{Conclusions}
In this paper, we presented \textbf{VDGS}, a visibility-aware 3D Gaussian Splatting framework for large-scale aerial scene reconstruction under highly imbalanced viewpoint distributions. By explicitly modeling the relationship between camera distribution and scene structure, VDGS addresses two critical challenges in large-scale 3DGS, namely suboptimal scene partitioning and insufficient supervision in low-visibility regions. Specifically, we introduced a visibility-guided partitioning strategy to achieve more balanced parallel optimization, together with a gradient compensation mechanism that enhances anchor densification in weakly observed areas, making VDGS an efficient and practical solution for real-world robotic applications such as large-scale mapping and autonomous driving.

\newpage
\bibliographystyle{IEEEtran}
\bibliography{refs}


\end{document}